\documentclass{article}
\usepackage[utf8]{inputenc}
\usepackage[backend=biber,style=numeric,sorting=nyt,maxbibnames=99]{biblatex}
\usepackage{lscape}
\usepackage{graphics}
\usepackage{adjustbox}
\usepackage{amsfonts}
\usepackage{graphicx}
\usepackage{changepage}
\usepackage{amsmath,bm}
\usepackage{lscape}
\usepackage{afterpage}
\usepackage{capt-of}
\usepackage{arxiv}

\usepackage[table]{xcolor}
\usepackage[capitalize]{cleveref}
\usepackage{csvsimple,booktabs}
\usepackage{tabularx}
\usepackage{adjustbox}
\usepackage{pgfplotstable,filecontents,booktabs}
\usepackage{multirow}
\usepackage{caption}
\usepackage{subcaption}
\usepackage{pgfplots}
\pgfplotsset{compat=1.8}
\usepgfplotslibrary{statistics}
\usepackage{comment}
\usepackage{pgfplotstable}
\usepackage[utf8]{inputenc}
\DeclareUnicodeCharacter{0301}{o}

\pgfplotstableset{
    color cells/.style={
        col sep=comma,
        string type,
         every nth row={6}{before row=\midrule},
        every nth row={10}{before row=\midrule},
every nth row={11}{before row=\midrule},
every head row/.style={before row={\toprule},after row=\midrule},
every last row/.style={after row={\toprule}},
        every first column/.style={ postproc cell content/.code={}, column type/.add={l}{}},
        postproc cell content/.code={%
                \pgfkeysalso{@cell content=\rule{0cm}{2.4ex}\cellcolor{black!##1}\pgfmathtruncatemacro\number{##1}\ifnum\number>50\color{white}\fi##1}%
                },
        columns/x/.style={
            column name={},
            postproc cell content/.code={}
        }
    }
}

\graphicspath{ {./Images/} }
\title{Disagreement amongst counterfactual explanations: How transparency can be deceptive}
\author{Dieter Brughmans, Lissa Melis and David Martens}

\begin{document}

\maketitle
\begin{abstract}
  
  Counterfactual explanations are increasingly used as an Explainable Artificial Intelligence (XAI) technique to provide stakeholders of complex machine learning algorithms with explanations for data-driven decisions. 
The popularity of counterfactual explanations resulted in a boom in the algorithms generating them. However, not every algorithm creates uniform explanations for the same instance. Even though in some contexts multiple possible explanations are beneficial, there are circumstances where diversity amongst counterfactual explanations results in a potential disagreement problem among stakeholders. Ethical issues arise when for example, malicious agents use this diversity to fairwash an unfair machine learning model by hiding sensitive features.
    As legislators worldwide tend to start including the right to explanations for data-driven, high-stakes decisions in their policies, these ethical issues should be understood and addressed.
    Our literature review on the disagreement problem in XAI reveals that this problem has never been empirically assessed for counterfactual explanations. Therefore, in this work, we conduct a large-scale empirical analysis, on 40 datasets, using 12 explanation-generating methods, for two black-box models, yielding over 192.0000 explanations. 
    Our study finds alarmingly high disagreement levels between the methods tested. A malicious user is able to both exclude and include desired features when multiple counterfactual explanations are available. This disagreement seems to be driven mainly by the dataset characteristics and the type of counterfactual algorithm. XAI centers on the transparency of algorithmic decision-making, but our analysis advocates for transparency about this self-proclaimed transparency. 
\end{abstract}

\section{Introduction}\label{sec:intro}
Artificial Intelligence (AI) or Machine Learning (ML) is rapidly evolving and disrupting various sectors, such as finance, healthcare, business (e.g., logistics, the labor market), education, and urban development. Besides the many benefits AI can create, multiple negative implications can be identified for each sector \cite{puavualoaia2023artificial}. One of the re-occurring challenges concerning AI is the need for transparency: many AI models are opaque and operate on a black-box basis, which makes it difficult - or sometimes impossible - to interpret and explain a decision that has been made. Therefore, Explainable Artificial Intelligence (XAI) has recently emerged as a much-needed research field. Next to an obvious focus on the predictability of AI models, model explainability is necessary for users, developers, and other stakeholders of real-life AI applications. Not only do people generally want to know an explanation for an algorithm-based decision, but also legislation is backing up this need. For example, in 2018 the European Union stated in the new General Data Protection Regulation (GDPR) that subjects of algorithmic decision-making are entitled to explanations. Users can ask for explanations of data-driven decisions that significantly influence their lives \cite{goodman2017european}.
Reaching a certain level of explainability in AI models is possible by either developing models that are inherently better interpretable - but sometimes have less predictive power - or by using post-hoc XAI techniques to generate explanations after predictions have been made with a black-box model. 

Even though seemingly good explanations for a model's decision can be generated by the use of a post-hoc XAI method, and consequently the model and its decisions are qualified as transparent, research on the \textit{uniformity} of these explanations is rather scarce. Many different post-hoc XAI methods exist and each method can generate different explanations for the same predicted outcome. Ergo, different stakeholders might be more interested in the explanations of one specific XAI method over another one. This raises the question of whether the transparency objective of XAI is achieved. In the literature, this phenomenon has been recently called the \textit{disagreement problem} \cite{krishna2022disagreement, neely2021order, roy2022don}, on which we will elaborate in \cref{sec:relatedwork}.  

\Textcite{miller2019explanation} takes knowledge from psychology, sociology, and cognitive sciences to identify what are "good" explanations. They argue that explanations are contrastive, selected, social and that probabilities most likely will not matter. The first means that people generally don't ask why a certain decision is made. People wonder why a certain decision is made \textit{instead} of another one. The second points to the fact that even though multiple explanations are possible to justify a decision, people are used to selecting one or two causes as \textit{the} explanation. The third means an explanation is always dependent on the beliefs of the user and the last refers to the preference of \textit{causes} over a probability or statistical relationship. These insights stress the usefulness of \textit{counterfactual (CF) explanations}, a post-hoc example-based XAI method which underlines a set of features that, when changed, alter a decision made by a model \cite{arrieta2020explainable}. Similarly to other post-hoc XAI techniques, different counterfactual methods might generate different explanations for the same instance and model, which raises ethical questions regarding the use of these techniques.

In this work, we will investigate the extent of the disagreement problem between popular counterfactual explanation methods. In \cref{sec:relatedwork}, we will situate counterfactual explanations in the diverse landscape of post-hoc XAI techniques and express how a lack of consistent evaluation methods for these techniques can lead to ambiguity in their explanations and ethical consequences. Consequently, in \cref{sec:exp}, we will quantify the disagreement amongst ten different counterfactual explanation methods next to Anchor and SHAP. The paper ends with conclusions and future research in \cref{sec:conclusion}.

\section{The diverse landscape of post-hoc explanations}\label{sec:relatedwork}

Post-hoc explanation methods are a subcategory of XAI that is concerned with explaining complex black-box models. In contrast to intrinsic explanation methods, they do not try to create interpretable white-box models, but are focused on explaining existing complex models \cite{linardatos2020explainable}. These methods are particularly interesting because their explanations seem to bypass the accuracy-explainability trade-off \cite{huysmans2006using}. This is a paradox stating that model performance often comes at a cost of model interpretability. However, these post-hoc methods are able to explain complex models and thus theoretically achieve both high performance and explainability at the same time. However, the quality of post-hoc explanation has often been a point of discussion \cite{fernandez2020explaining, doshi2017towards}.

The difficulty concerning these methods is their evaluation. Because the model is not intrinsically explainable, it is difficult to assess the quality of such  explanations. The field of XAI evaluation has come up with different metrics to quantify this quality, however, no consensus has currently been reached. Since we cannot strictly quantify the quality of a post-hoc explanation method, many methods are proposed and used. This has led to ambiguity amongst explanations: explanations for the same instance are different depending on the post-hoc explanation method used (the disagreement problem). The quantified lack of uniformity in explanations has already been investigated for several post-hoc explanation methods \cite{krishna2022disagreement,neely2021order,roy2022don}, however, to the best of our knowledge, this problem has not yet been investigated for counterfactual explanations, which is the main contribution of this work.


We first give an overview and classification of the post-hoc explanation methods used for comparison in this work in \cref{sec:overviewposthoc}. In \cref{sec:evaluationposthoc} we discuss how post-hoc explanation methods are currently evaluated and address some core issues regarding this topic. Lastly, \cref{sec:disagreement} elaborates on the existing research on the disagreement problem and the need to apply this research to counterfactual explanations.

\subsection{Counterfactual explanations and recent post-hoc explanation methods} \label{sec:overviewposthoc}

\begin{figure}[hhtp]
     \centering
     \includegraphics[scale=0.60]{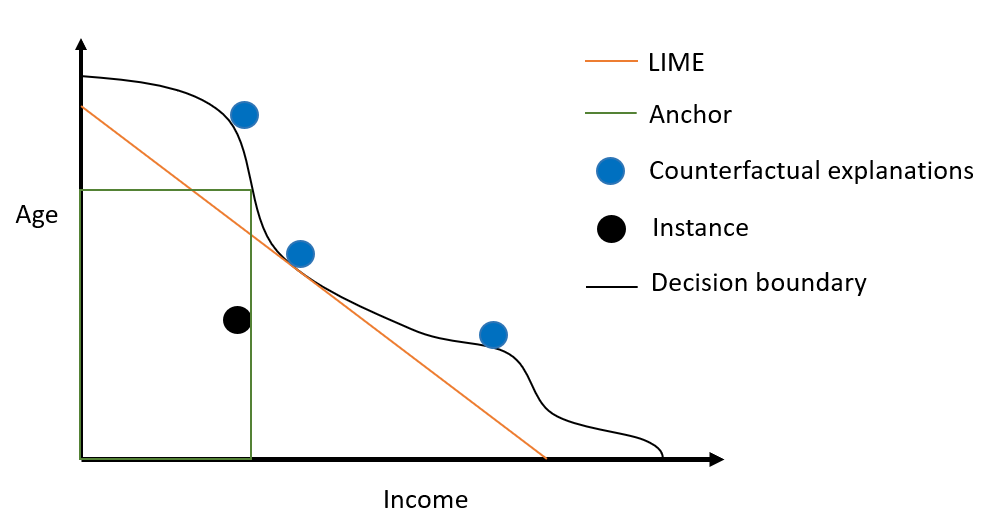}
     \caption{Figurative toy example of LIME, Anchors and counterfactual explanations for a loan approval predictive model, based on \textcite{brughmans2021nice}, \textcite{molnar2018guide} and \textcite{ribeiro2016should}}
     \label{fig:example}
 \end{figure}

Post-hoc explanations are techniques to explain AI decisions made after the model has been trained. The most popular methods can be divided into two groups: feature-based techniques (also called attribution methods) and example-based (also called instance-based) techniques \cite{dwivedi2023explainable, molnar2018guide}. The first group contains methods like local interpretable model-agnostic explanations (LIME), Shapley additive explanations (SHAP), and other feature importance techniques. The second group, example-based post-hoc explanation methods, contains Anchors and counterfactuals. We will briefly explain the methods used in our experiments: SHAP, Anchors, and counterfactual explanations. \Cref{fig:example} provides a figurative example of the different XAI methods to explain why a person (the instance) is predicted not to get their loan approved. For a more detailed description and examples, we refer to \textcite{molnar2018guide} and the works mentioned below.

 \paragraph{(LIME and) SHAP} SHAP \cite{lundberg2017unified} and LIME  \cite{ribeiro2016should} are similar in the sense that the impact of a certain feature is measured related to the predictive outcome. The basic idea of LIME is to sample instances in the neighborhood of the instance that is given to the ML model and then train an interpretable model like linear regression or decision tree to explain this neighborhood. The interpretable model can consequently be used to explain the prediction that is made by the actual, black-box model. For tabular data, which are used in the experiments of this work, the issue is to define the neighborhood of an instance. If LIME would only sample closely around the given instance, chances are high all predictions will be exactly the same and LIME cannot comprehend how predictions change. Therefore, samples are taken broadly, e.g., by using a normal distribution. A major disadvantage of LIME is that the explanations differ depending on the samples used, which makes the explanations unstable and manipulable. Therefore, we do not include LIME in our comparisons.
 
 A better feature-based technique can be found in SHAP, which combines the locality of LIME with the concept of Shapley values from coalitional or cooperative game theory. The contribution of each feature (player) to the prediction that is made by the model (outcome of the game) for a given instance is calculated. Moreover, the contribution of cooperation between players (multiple features) is examined. The average marginal contribution of a feature value across all cooperations is called the shapley value\footnote{A common misinterpretation of the Shapley value is that it amounts to the difference in prediction after removing the feature from the model training.}. Because there are $2^k$ possible cooperations, for which models need to be trained, calculating all the Shapley values is computationally expensive. Therefore, by using the LIME-inspired sampling, the SHAP algorithm decreases the computation time. 

 \paragraph{Anchors} Anchors or scoped rules \cite{ribeiro2018anchors} are high-precision easy-to-understand if-then rules. They portray feature conditions together with a predictive outcome. The rules are called Anchors because any changes to other features than the ones mentioned, will not result in another prediction. In contrast to e.g., LIME, Anchors will provide a region of instances to describe the model's behavior. They are consequently less instance-specific. For example, imagine if a person applies for a loan at a bank. This person is 50 years old, has a monthly income of \$2000, his gender is male and he currently has \$5000 in debt. A model has predicted that the loan application should be declined. The corresponding Anchor could then be: if the monthly income is lower than \$5000 and the age is higher than 35, then predict that the loan application would be accepted.

 \paragraph{Counterfactual explanations} Counterfactual explanations describe a combination of feature changes that would alter the predicted class \cite{SEDC}. In other words, they determine what features should change in order to change the prediction and are consequently sometimes called what-if statements. As mentioned in \cref{sec:intro}, this type of explanation is especially human-friendly because they are contrastive and selective \cite{miller2019explanation}. Counterfactual explanations are somewhat the opposite of Anchors. To revisit the same example: the person asking for a loan wants to know why he will not get one. A counterfactual explanation could then be: if your monthly income rises to \$5000, you will get a loan. Besides its human-friendly nature, another advantage of the counterfactual method is that it does not need access to the data or the model itself. Only the model’s prediction function is required to generate the explanations. 
 Because of their many benefits, \emph{many} different counterfactual methods came into existence. \textcite{guidotti2022counterfactual} and \textcite{verma2020counterfactual} give an overview of counterfactual explanation techniques, however, to date, the state-of-the-art further unfolded with e.g., the introduction of NICE, a counterfactual generation algorithm which simultaneously achieves 100\% coverage, model-agnosticism and fast counterfactual generation for different types of classification models. 

 Feature importance estimates can be biased because it is dependent on the executed sampling \cite{fernandez2020explaining}, while example-based methods like counterfactuals have no additional assumptions or mysteries at the back end of the method. They just tell the user how to change the prediction. On the downside, it is arguable if one instance can fully capture the complexity of a model. Furthermore, the literature recently sees a sprawl of different counterfactual methods, which \emph{possibly} leads to different respective explanations. The question rises whether these counterfactual methods are still useful for stakeholders and which specific post-hoc explanation method to choose. To answer these questions, the field of XAI evaluation has tried to evaluate the quality of post-hoc explanation methods.

\subsection{Ambiguity due to a lack of consistent evaluation metrics for post-hoc explanations}
\label{sec:evaluationposthoc}

As referred to in \cref{sec:intro} and \cref{sec:overviewposthoc}, evaluating XAI methods is a research field in its infancy today, even though a strong need for evaluation methods is identified by multiple authors such as \textcite{rosenfeld2021better}. 
One reason for the limited amount of research done in this field can be the simple fact that evaluating XAI methods is \textit{difficult}, especially for post-hoc explanation methods. These are designed to explain complex models with a black-box nature. By definition, we don't know the logic involved in a decision made by such models, thus we cannot compare the output of a post-hoc explanation method to a model's ground truth. Some have tried generating synthetic datasets for which the ground truth is known \cite{liu2021synthetic, yalcin2021evaluating, amiri2020data}. However, this does not completely solve the problem. First of all, the comparison of this ground truth with the explanations is not always straightforward. This results in multiple metrics, for example, \textcite{liu2021synthetic} look at faithfulness, monotonicity, ROAR, GT-Shapley, and infidelity. Each of these metrics has its own pros and cons. Consequently, the best metric is subjective and application-specific \cite{adebayo2018sanity}. Secondly, a model might not learn all relationships in the data correctly and show bias. The goal of XAI is to explain the model and not the data. In other words, the model's ground truth might be different from the data's ground truth and therefore comparing model explanations with the data's ground truth can be inaccurate. Lastly, \textcite{liu2021synthetic} agree with an existing gap between synthetic and real-life datasets. They also cite that using their synthetic datasets and connected ground truths cannot be seen as a replacement for human-centered evaluation of XAI techniques, which is a mostly qualitative way of evaluating explanations.

\textcite{vilone2021notions} divide XAI evaluation techniques into two groups: those that involve  human-centred evaluations and those that evaluate with objective metrics. The first require human participants to give qualitative or quantitative feedback to XAI explanations, typically through surveys. For the second, to this day, more than 35 metrics have been proposed in the literature to evaluate XAI explanations. Examples of these metrics are, among others, actionability (knowledge is useful to the end-user), efficiency (computational speed of the algorithm), simplification (minimal features), stability (similar instances should provide similar explanations), etc. The authors conclude that the boom in the amount of evaluation metrics calls for a general consensus among researchers on how an explanation should be evaluated. 

Note that these objective metrics are sometimes hard to quantify. Qualitative quality properties are therefore often quantified in numbers. For counterfactual explanations, popular properties are proximity, sparsity, and plausibility \cite{verma2020counterfactual}. Proximity is a property that is somehow used in every counterfactual algorithm. It tries to measure the total change that is suggested by the counterfactual explanations with a distance metric \cite{CFproto, mothilal2020explaining, wexler2019if}. It is intuitive that less change is better than more change in most situations. Sparsity is a special case of proximity. It refers to the number of features in the explanation \cite{karimi2020model,dandl2020multi,laugel2018comparison}. The argument is that shorter explanations are more comprehensible for humans than longer ones \cite{miller1956magical}. Finally, plausibility is a more conceptual property that refers to the closeness to the data manifold \cite{pawelczyk2020counterfactual}. For example, in a credit scoring context, advising someone to wait 200 years in order to get a loan, is not plausible. 

Counterfactual explanations have an additional advantage in comparison to feature importance methods. The latter estimate the influence of each feature on the predicted score. These estimates potentially suffer from bias and features which have almost no influence on the model's decision might be labeled important \cite{fernandez2020explaining}. Counterfactual explanations don't suffer from this bias, applying the suggested changes of a counterfactual explanation will always lead to a change in prediction. Consequently, a counterfactual explanation is always "correct" in one sense. However, this does not guarantee that these are quality explanations.

The ambiguity of measuring the quality of counterfactual explanations has led to the development of many counterfactual algorithms and possibly as many different explanations \cite{verma2020counterfactual, guidotti2022counterfactual}. As a result, when a stakeholder now wants to use counterfactual explanations, he is presented with many options. This might be an advantage or can lead to the disagreement problem.

\subsection{The disagreement "problem"} \label{sec:disagreement}

The disagreement problem in explainable AI arises when different interpretability methods, used to explain a given AI model, produce conflicting or contradictory explanations. Because of a lack of broadly used evaluation methods, this is often the case, resulting in explanations that are generally non-consistent and thus ambiguous. \textcite{neely2021order} raise the question of whether agreement as an evaluation method for XAI methods is suitable. When assuming agreement as an evaluation method, low agreement would mean only a few of the XAI methods are right, while the others are far from ideal. However, low agreement is not necessarily a bad thing.

Ambiguity can actually be valuable or result in possible ethical consequences \cite{martens2022data}. It all depends on the context in which XAI methods are used \cite{bordt2022post}. \textcite{mothilal2020explaining} argues that diversity among counterfactual explanations is beneficial. This for one increases the chance of generating usable explanations. For example, when someone is not allowed to get a loan according to an ML model, and the only counterfactual explanation is to change their sex or lower their level of education, this explanation is not useful. People prefer to get an actionable explanation, such as 'increase your income with \$X'. This actionability is not uniform over all decision subjects. Therefore, providing multiple explanations increases the chance that one explanation is useful for this specific user. 

\textcite{bordt2022post} examine when ambiguity in explanations is problematic. They differentiate between a cooperative and adversarial context. In a cooperative context, all stakeholders have the same interests. For example, in most medical applications of AI, both doctors and patients have the same goal: to improve or manage the patient's health. In adversarial contexts, this is not the case. Here, different parties have opposite interests. For example, when a student is denied admission to a prestigious university, the student is interested in challenging this decision. Another example is an autonomous car crashing into a wall to avoid a pedestrian. Insurance companies have other interests than the owner of the car or the developers of the software that steers the car's driving decisions. Another example is a denied bank loan: the bank and the client have different interests. In these cases, it might not be in the model user's best interest to look for the most correct or elaborate explanation of a decision that is made. The model user will most likely choose the explanation that fits their best interest, \textit{if} diverse explanations are available. 
An adversarial context can lead to all kinds of ethical issues \cite{martens2022data}. \textcite{Aïvodji} examine the use of post-hoc explanations to fairwash or rationalize decisions made by an unfair ML model, while \textcite{slack2020fooling} and \textcite{lakkaraju2020fool} investigate the discriminatory characteristics of explanations. Imagine a model using a prohibited feature such as e.g., gender or race, or a feature that is linked to one of these e.g., zip code, but other more neat explanations are available. The model user could choose to ignore the discriminatory explanations and use another one instead. When considering the ethical consequences of disagreement, consensus amongst explanations might be desired. Therefore, consensus between explanations could be seen as a training objective to increase user trust \cite{reckoning2023, hinnsUnderspec}. Namely, if two explanations are consensual, the ethical consequences of choosing one XAI method over another one are less severe.


\subsubsection{Related work} However, these ethical issues can only arise if there actually is ambiguity. The first one to measure the disagreement problem in XAI is \textcite{neely2021order}. They compare LIME, Integrated Gradients, DeepLIFT, Grad-SHAP, Deep-SHAP, and attention-based explanations with a rank correlation (Kendall’s $\tau$) metric. They conclude there is only low agreement in the explanations of these methods. \textcite{krishna2022disagreement} expand the previous study by comparing LIME, KernelSHAP, Vanilla Gradient, Gradient Input, Integrated Gradients, and SmoothGrad, once again finding disagreement amongst explanation of different methods, especially when the model complexity increases. Instead of only using a rank correlation metric, they use a feature agreement, (signed) rank agreement, sign agreement, and rank correlation. Depending on the type of data (tabular, text or image data), they use different of the above-mentioned evaluation metrics. Next to a quantitative comparison, the authors also perform a qualitative study on how practitioners handle the disagreement problem. 84\% of practitioners interviewed by \textcite{krishna2022disagreement} mentioned encountering the disagreement problem on a day-to-day basis. They report there is no principle evaluation method to decide on which explanations to use, therefore, they simply choose to generate explanations with the XAI method they are most familiar with.
\textcite{han2022explanation} extend the study of \textcite{krishna2022disagreement} to investigate why the disagreement problem exists for these methods. They conclude that different XAI methods approximate a black-box model over different neighborhoods by applying other loss functions. If two explanations are trained to predict different sets of perturbations, then the explanations are each accurate in their own domain and may disagree. 
A more focused disagreement problem study can be found in \textcite{roy2022don} where the explanations of LIME and SHAP are investigated for one single defect prediction model. They calculate the feature, rank, and sign agreement also proposed by \textcite{krishna2022disagreement}. They conclude that LIME and SHAP disagree more on the ranking of important features compared to the sign of their importance.

\begin{table}[hhtp]
    \centering
    \begin{adjustbox}{max totalsize={1\textwidth}{1\textheight},center}
    \begin{tabular}{lclccr}
    \toprule
        Authors & Nb. of  & Type of  & Nb. of  & Nb. of XAI  &  XAI  \\
        &datasets&datasets&models&methods&\\
        \midrule
        \textcite{neely2021order} & 5 & Text & 2 & 6 & LIME, Integrated Gradients, DeepLIFT, \\
        \multicolumn{5}{c}{} & Grad-SHAP, Deep-SHAP, and attention-based explanations \\
        \textcite{krishna2022disagreement} & 4 & Tabular, Text and Image & 8 & 6 & LIME, Kernel SHAP and Integrated Gradients\\
        \textcite{roy2022don} & 4 & Tabular & 1 & 2 & LIME, SHAP\\
       This work & 40 & Tabular & 2 & 12 & SHAP, Counterfactuals and Anchor\\
        \bottomrule
    \end{tabular}
    \end{adjustbox}
    \caption{Literature overview of the quantitative evaluation of the disagreement between post-hoc XAI methods}
    \label{tab:litoverview}
    
\end{table}

\Cref{tab:litoverview} gives an overview of the scarce literature on the quantitative evaluation of disagreement between XAI methods relative to our work. To the best of our knowledge, the disagreement problem has not yet been quantified for counterfactual XAI methods. This is remarkable because recently there has been a boom in the number of such algorithms. \textcite{guidotti2022counterfactual} identify 60 unique counterfactual algorithms in their recent survey. In this manuscript, we study the disagreement problem between counterfactual explanations for tabular data. We investigate disagreement between 10 selected counterfactual algorithms but also the disagreement with different post-hoc explanation methods such as SHAP \cite{lundberg2017unified} and Anchors \cite{ribeiro2018anchors}. Moreover, we extend the size of our study to 40 datasets (instead of 4-5) in order to confidently make more general conclusions. 

 It should be noted that morally the disagreement problem for counterfactual explanations is similar to the Rashomon effect, introduced by \textcite{hasan2022mitigating}. This effect concerns the diversity in multiple counterfactual explanations generated by \emph{the same} counterfactual algorithm for the same instance and classifier. One explanation might say to change feature A (e.g., wait 5 years in order to get a loan), while another might say to change feature B while not adapting A (e.g., make sure your income increases with \$500 in order to get a loan immediately). This initially seems like a contradiction as well. For example, the DiCE algorithm is focused on generating multiple explanations \cite{mothilal2020explaining}. In contrast to the Rashomon effect, the disagreement problem investigates diversity amongst \emph{different} counterfactual explanation algorithms for the same instance and classifier. However, at the heart of the matter, the Rashomon effect and the disagreement problem face the same ethical issues and moral hazards: who chooses which explanation will be used?

\section{The quantified disagreement amongst counterfactual explanation methods}
\label{sec:exp}
In this section, we aim to quantify the disagreement between counterfactual explanation methods. We first illustrate the problem and research questions with an example in \cref{sec:example}. \Cref{sec:setup} clarifies the large-scale experimental setup. \Cref{sec:results} answers the research questions by providing metrics to quantify the disagreement amongst counterfactual algorithms and using these metrics in our large-scale experimental setup. Note that measuring the disagreement between counterfactual explanations comes with some new challenges. First of all, counterfactual explanations provide a set of features without ranking them. This makes measures such as (signed) rank agreement useless. Second, counterfactual explanations consist of variable sizes. Some algorithms might suggest six feature changes while others might only suggest two.

\subsection{Example} \label{sec:example}
\Cref{tab:exampleinstance} illustrates the disagreement problem for an example instance retrieved from the Adult dataset \cite{Dua:2019}. This dataset can be used to predict if a person would have an annual income higher or lower than \$50K. The person depicted in the instances is predicted to have an income lower than \$50K. Consequently, ten different counterfactual explanation algorithms generated counterfactual instances in order to tell which features should change in the original instance in order to change the prediction. 

Firstly, it should be noted that one of the counterfactual explanation methods, CBR, is not able to find a counterfactual instance for the given original instance, while the others do find one. When having the need to explain a certain instance, it is consequently useful that other explanation methods are able to find explanations and thus, that some disagreement amongst methods is existing. However, in an adversarial context, a malicious counterfactual generating user, wishing to avoid a certain feature e.g., sex or race, is able to do so by simply selecting a counterfactual method that does not include these features, i.e., wants to change these features with respect to the original instance, such as DiCE, NICE(plaus) or NICE(spars). Imagine we are predicting whether or not this person would be qualified to get a loan from a bank. An ML model that uses features like sex or race would then be unethical and discriminatory. The decision maker would be able to \textit{hide} this fact by secretly choosing a counterfactual method that does not include sex or race in the explanations. This way, the unfair ML model can still be "rationally" explained. Vice versa, if the malicious user explicitly wants to \textit{include} a certain feature in an explanation, e.g., hours per week, they can do so with CFproto, NICE(none) or NICE(plaus). And this once again by simply choosing among the diverse explanations, without any need to impose constraints on the counterfactual generating search. This shows the arbitrariness/disagreement of the methods and the power that it brings to the user of counterfactual generating methods. A user can include, as well as avoid, almost any desired feature in the given explanation.

This example clearly illustrates the possible existence of the disagreement problem and the  ethical consequences resulting from this existence. In the following sections, we examine how easy it is to abuse the disagreement problem by malicious agents and the driving factors that cause the disagreement problem in the first place.

\begin{table}[hhtp]
    \centering
     \begin{adjustbox}{max totalsize={1\textwidth}{0.65\textheight},center}
    \begin{tabular}{lcccccccccccccc}
     \toprule
	&	Workclass	&	Marital	&	Relationship	&	Race	&	Sex	&	Age	&	Fnlwgt	&	Edu.	&	Edu. 	&	Occupation	&	Capital	&	Capital	&	Hours	&	Native 	\\
	&		&	 status	&		&		&		&		&		&		&	num	&		&	gain	&	loss	&	\textbackslash week	&	country	\\
\midrule
\midrule
Instance	&	Local gov.	&	Wid.	&	Other relative	&	Native A/A	&	F	&	26	&	195693	&	1st-4th	&	9	&	Craft-repair	&	0	&	0	&	40	&	France	\\
\midrule
\midrule
CBR	&		&		&		&		&		&		&		&		&		&		&		&		&		&		\\
CFproto	&	\cellcolor[gray]{0.8}Self empl.	&	Wid.	&	\cellcolor[gray]{0.8}Own-child	&	Native A/A	&	\cellcolor[gray]{0.8}M 	&	\cellcolor[gray]{0.8}44	&	\cellcolor[gray]{0.8}90688	&	\cellcolor[gray]{0.8}Doctorate	&	\cellcolor[gray]{0.8}15	&	\cellcolor[gray]{0.8}Other service	&	0	&	0	&	\cellcolor[gray]{0.8}51	&	\cellcolor[gray]{0.8}Portugal	\\
WIT	&	Local gov.	&	\cellcolor[gray]{0.8}Never marr.	&	\cellcolor[gray]{0.8}Unmarried	&	\cellcolor[gray]{0.8}Black	&	F	&	\cellcolor[gray]{0.8}29	&	\cellcolor[gray]{0.8}197932	&	1st-4th	&	9	&	\cellcolor[gray]{0.8}Handlers-cleaners	&	0	&	0	&	40	&	\cellcolor[gray]{0.8}Ecuador	\\
GeCo	&	Local gov.	&	\cellcolor[gray]{0.8}Marr.	&	Other relative	&	\cellcolor[gray]{0.8}Black	&	F	&	\cellcolor[gray]{0.8}17	&	\cellcolor[gray]{0.8}195695	&	1st-4th	&	9	&	Craft-repair	&	0	&	0	&	40	&	\cellcolor[gray]{0.8}Hong K.	\\
NICE(none)	&	Local gov.	&	Wid.	&	Other relative	&	\cellcolor[gray]{0.8}Black	&	F	&	\cellcolor[gray]{0.8}43	&	\cellcolor[gray]{0.8}112763	&	1st-4th	&	9	&	Craft-repair	&	\cellcolor[gray]{0.8}8614	&	0	&	\cellcolor[gray]{0.8}43	&	\cellcolor[gray]{0.8}Hong K.	\\
NICE(plaus)	&	Local gov.	&	Wid.	&	Other relative	&	\cellcolor[gray]{0.8}Black	&	F	&	\cellcolor[gray]{0.8}43	&	\cellcolor[gray]{0.8}112763	&	1st-4th	&	9	&	Craft-repair	&	\cellcolor[gray]{0.8}8614	&	0	&	\cellcolor[gray]{0.8}43	&	\cellcolor[gray]{0.8}Hong K.	\\
\midrule
DiCE	&	Local gov.	&	Wid.	&	Other relative	&	Native A/A	&	F	&	\cellcolor[gray]{0.8}50	&	195693	&	1st-4th	&	9	&	Craft-repair	&	\cellcolor[gray]{0.8}28533	&	0	&	40	&	France	\\
NICE(prox)	&	Local gov.	&	Wid.	&	Other relative	&	Native A/A	&	F	&	26	&	195693	&	1st-4th	&	9	&	Craft-repair	&	\cellcolor[gray]{0.8}8614	&	0	&	40	&	France	\\
NICE(spars)	&	Local gov.	&	Wid.	&	Other relative	&	Native A/A	&	F	&	26	&	195693	&	1st-4th	&	9	&	Craft-repair	&	\cellcolor[gray]{0.8}8614	&	0	&	40	&	France	\\
SEDC	&	Local gov.	&	\cellcolor[gray]{0.8}Never marr.	&	\cellcolor[gray]{0.8}Wife	&	Native A/A	&	\cellcolor[gray]{0.8}M 	&	\cellcolor[gray]{0.8}39	&	195693	&	\cellcolor[gray]{0.8}Masters	&	\cellcolor[gray]{0.8}10	&	Craft-repair	&	0	&	0	&	40	&	France	\\
\bottomrule															
    \end{tabular}
    \end{adjustbox}
    \vspace{3mm}
   \\ \scriptsize(Edu. = Education, Local gov. = Local government, Wid. = Widowed, Native A/A = Native Alaskan/American, Self empl. = Self employed, Never marr. = Never Married, Marr. = Married, Hong K. = Hong Kong)
    \vspace{3mm}
    \caption{Example instance (Adult dataset)}
    \label{tab:exampleinstance}
\end{table}

\subsection{Experimental setup} \label{sec:setup}
\Cref{tab:TableDataSummary} gives an overview of the 40 tabular datasets we use for our study. A test set is created for each dataset by comprising 20\% of the data with a minimum of 200 instances. This means that e.g., for the threeOf9 dataset, we do not use 102 instances in the test set, but we use 200. The remaining data is used as the training set for training a Random Forest classifier (RF) and an Artificial Neural Network (ANN). The final two columns of \cref{tab:TableDataSummary} display the AUC values obtained for both classifiers. The hyper-parameters of both models are trained using a five-fold cross-validation approach. Subsequently, we generate counterfactual explanations using all algorithms for a random sample of 200 instances from the test set. In total, we generate 200 counterfactual explanations for 10 counterfactual algorithms, Anchors, and SHAP for 2 classifiers on 40 datasets, resulting in a sample size of 192,000 explanations.

\begin{table}[hhtp]
\footnotesize
\centering
   \begin{adjustbox}{max totalsize={1\textwidth}{0.65\textheight},center}
		\begin{tabular}{lrrrrccccc}
  \toprule
		\textbf{Name} & \textbf{\#Inst.}  & \textbf{\#Feat.} & \textbf{\#Cat.}& \textbf{\#Num.}& \textbf{Class}& \textbf{AUC} & \textbf{AUC}\\
		&  &  & \textbf{feat.} & \textbf{feat.} & \textbf{imbalance} & \textbf{(ANN)} & \textbf{(RF)}\\
			\midrule
            adult	&48,842&	14&	5	&9&	0.761&	0.903&	0.913\\
            agaricus\_lepiota&	8154&	22	&21	&1	&0.481&	1.000&	1.000\\
            australian	&690&	14&	7&	7&	0.445&	0.905&	0.940\\
            breast\_w&	699	&9&	8&	1&	0.345&	0.991&	0.997\\
            buggyCrx&	690&	15&	8&	7&	0.555&	0.921&	0.949\\
            chess&	3196&	36&	36&	0&	0.522&	1.000&	0.999\\
            churn&	5000&	20&	4&	16&	0.142&	0.872&	0.919\\
            clean2&	6598&	168	&0	&168&	0.154&	1.000&	1.000\\
            coil2000&	9822&	85&	84&	1&	0.060&	0.691&	0.745\\
            credit\_a&	690&	15&	8&	7&	0.555&	0.902&	0.910\\
            credit\_g&	1000&	20&	17&	3&	0.700&	0.663&	0.731\\
            crx&	690&	15&	8&	7&	0.445&	0.859&	0.941\\
            diabetes&	768&	8&	0&	8&	0.349&	0.823&	0.851\\
            dis&	3772&	29&	23&	6&	0.985&	0.895&	0.989\\
			GAMETES1&	1600&	20&	20&	0&	0.500&	0.636&	0.648\\
			GAMETES2&	1600&	20&	20&	0&	0.500&	0.746&	0.780\\
            GAMETES3&	1600&	20&	20&	0&	0.500&	0.664&	0.722\\
            GAMETES4&	1600&	20&	20&	0&	0.500&	0.690&	0.705\\
            german&	1000&	20&	17&	3&	0.700&	0.718&	0.758\\
            Hill\_Valley	&1212	&100	&0&	100&	0.505&	0.993&	0.557\\
            hypothyroid&	3163&	25&	18&	7&	0.952&	0.975&	0.988\\
            kr\_vs\_kp&	3196&	36&	36&	0&	0.522&	1.000&	0.999\\
            magic&	19,020&	10&	0&	10&	0.352&	0.922&	0.937\\
            mofn\_3\_7\_10&	1324&	10&	10&	0&	0.779&	1.000&	1.000\\
            monk1&	556&	6&	6&	0&	0.500&	1.000&	1.000\\
            monk2&	601&	6&	6&	0&	0.342&	1.000&	0.896\\
            monk3&	554&	6&	6&	0&	0.520&	0.992&	0.986\\
            mushroom&	8124&	22&	21&	1&	0.482&	1.000&	1.000\\
            parity5+5&	1124&	10&	10&	0&	0.504&	1.000&	0.674\\
            phoneme&	5404&	5&	0&	5&	0.294&	0.906&	0.970\\
            pima&	768&	8&	0&	8&	0.349&	0.867&	0.819\\
            profb&	672&	9&	3&	6&	0.333&	0.633&	0.676\\
            ring&	7400&	20&	0&	20&	0.505&	0.990&	0.992\\
            spambase&	4601&	57&	0&	57&	0.394&	0.974&	0.988\\
            threeOf9&	512&	9&	9&	0&	0.465&	0.972&	0.999\\
            tic\_tac\_toe&	958&	9&	9&	0&	0.653&	0.997&	1.000\\
            tokyo1&	959&	44&	2&	42&	0.639&	0.962&	0.983\\
            twonorm&	7400&	20&	0	&20	&0.500&	0.996&	0.997\\
            wdbc&	569&	30&	0&	30&	0.371&	0.973&	0.981\\
            xd6&	973&	9&	9&	0&	0.331&	1.000&	1.000\\
            \bottomrule
		\end{tabular}
  \end{adjustbox}

 \caption{Descriptive statistics and performance metrics of all 40 binary datasets.}
\label{tab:TableDataSummary}
\end{table}

\begin{table}[hhtp]
\footnotesize
\centering
	\begin{tabular}{llccc}
 \toprule
	    
		Name & Author & Spars & Prox & Plaus \\
		\midrule
  CBR  &\textcite{keane2020good}& x& &x  \\
  CFproto &\textcite{CFproto} & x & x& x \\
		WIT  &\textcite{wexler2019if}& & x &x  \\
  GeCo &\cite{schleich2021}& x & x &x \\
		NICE (none)&\textcite{brughmans2021nice} &  & x & x   \\
   NICE (plaus) &\textcite{brughmans2021nice}& x &  & x   \\
   \midrule
        DiCE  &\textcite{mothilal2020explaining}& & x &   \\
        NICE (prox) &\textcite{brughmans2021nice}&   & x &    \\
         NICE (spars) &\textcite{brughmans2021nice}& x &  &    \\
         SEDC &\textcite{SEDC}& x & &  \\

  \bottomrule
	\end{tabular}
\caption{Overview of the counterfactual algorithms used for comparison.}
\label{tab:tableAlgoOverview}
\end{table}

 We selected a total of 12 post-hoc explanation methods to study the disagreement problem. We focus on ten counterfactual algorithms which are suited for tabular data, are model-agnostic and have their code publicly available. They are depicted in \cref{tab:tableAlgoOverview}. The final selection includes the following counterfactual algorithms: DiCe \cite{mothilal2020explaining}, CFproto \cite{CFproto}, WIT \cite{wexler2019if}, CBR \cite{keane2020good}, SEDC \cite{fernandez2020explaining}, GeCo \cite{schleich2021} and four types of the NICE algorithm \cite{brughmans2021nice} to investigate the uniformity of their explanations. We refer to their respective manuscripts for detailed descriptions of the different counterfactual algorithms. Moreover, we also look at their disagreement with both SHAP and Anchors. As we mentioned in \cref{sec:evaluationposthoc}, there is no consensus on what defines the quality of a counterfactual explanation, which resulted in many algorithms optimizing explanations for different evaluation metrics. When we compare algorithms optimized or evaluated for other metrics, some form of disagreement is expected. However, when comparing explanations from algorithms that optimize for the same metric, one might expect less disagreement. We notice two distinct groups in \cref{tab:tableAlgoOverview} (divided by a horizontal line). The first six algorithms optimize for plausibility and the last four do not. The second group is only interested in providing counterfactual instances that are close to the original instance and thus called \texttt{Prox}\footnote{We take algorithms that optimize for sparsity and proximity together because sparsity is a special case of proximity, as is explained in \cref{sec:evaluationposthoc}.}. The first group is called \texttt{Plaus}. 

For the counterfactual explanations, we consider a feature to be present in the explanation if the counterfactual instance indicates to change the feature compared to the original instance. For Anchors we consider a feature present simply when it is mentioned in the Anchor explanation. For SHAP we take into account the seven most important features as features present in an explanation (based on \textcite{miller1956magical}). 

\subsection{Results} \label{sec:results}

\subsubsection{To what extent can counterfactual disagreement be abused by malicious agents?}\label{sec:question1}
The main issue with disagreement amongst counterfactual explanations is that malicious users can select certain explanations to rationalize decisions made by unfair or discriminating models. This can be done by either avoiding certain features to convince stakeholders that they are irrelevant or the other way around, by including certain features to insinuate that they are the main driver of the decision-making process.

We first check, how easy it is to \emph{exclude} a certain feature from a counterfactual explanation. This can be done by looking at the percentage of features that are not present in at least one explanation. \Cref{eq:RFeatureExcl} formalizes this metric which we call relative feature exclusion. In this metric, the numerator counts the unique features that are \emph{not} present in the explanations of certain methods $a$ to $n$. This number is divided by the total number of features $F_D$ in a dataset $D$. \Cref{tab:exclrf} and \cref{tab:exclann} show the average relative feature exclusions for different datasets, XAI-methods and classifiers.

\begin{equation}
\label{eq:RFeatureExcl}
\textrm{Relative feature exclusion}_{[a,n]} = \frac{|(F_D \setminus E_a) \cup (F_D \setminus E_b)  ... \cup (F_D \setminus E_n)|}{|F_D|}
\end{equation}

Next, we investigate the possibility to \emph{include} a random feature into a counterfactual explanation. For this, we introduce a metric called relative feature span, see \cref{eq:RFeatureSpan}. It measures the percentage of all features that is present in at least one explanation. The numerator equals the absolute feature span and measures the size of the union of all explanations of all explanation methods $a$ to $n$ in the comparison. The absolute feature span divided by $F_D$ is the relative feature span. A higher feature span most likely results from a higher disagreement amongst methods. Consequently, the user will be able to choose many features as part of the explanation. The maximum relative feature span of 1 is achieved when every single feature is used in at least one explanation. These relative feature spans are shown in \cref{tab:spanrf} and \cref{tab:spanann}.

\begin{equation}
\label{eq:RFeatureSpan}
\textrm{Relative feature span}_{[a,n]} = \frac{|E_a \cup E_b  ... \cup E_n|}{|F_D|}
\end{equation}

If we revisit the example of \cref{sec:example} and assume that the user only has the first two counterfactual explanation algorithms, CFproto and WIT, available. 
The relative feature exclusion between these two methods amounts $\frac{|(F_D \setminus E_{CFproto}) \cup (F_D \setminus E_{WIT})|}{|F_{Adult}|}$ or 57.1\%, meaning that 57.1\% of the features can be avoided by the user when using only these two methods. 
The relative feature span of both methods amounts $\frac{|E_{CFproto} \cup E_{WIT|}}{|E_{Adult}|}$ or 85.7\%. This means that 85.7\% of the features are present in the explanations of CFproto and WIT, and can consequently be chosen by the user. If all counterfactual explanation methods of \cref{tab:exampleinstance} are available to the user the overall relative feature exclusion equals 100\%. This means that any of the features can be chosen to be left out of the explanation if all methods are available to the user. The overall relative feature span equals 92.9\%. Only the feature 'capital loss' is never used in the explanations. 

Our results in \cref{tab:exclrf} and \ref{tab:exclann} show that excluding certain features is particularly easy when multiple explanations are available. The average relative feature exclusion is over 99.6\% for both classifiers over all counterfactual methods, and 99.8\% if we include Anchors and SHAP. For many datasets, the average relative feature exclusion is even 100.0\%. Meaning that for every instance that has to be explained, every feature of choice can be excluded from the explanation. These results show that it is fairly easy to avoid sensitive features in order to falsely justify model decisions.

\begin{minipage}{0.5\textwidth}
\begin{adjustbox}{max totalsize={0.95\textwidth}{1\textheight},center}
\begin{tabular}{lcccc}
\toprule
Dataset	&	Prox	&	Plaus	&	All CF	&	All	\\
\midrule
adult	&	97.8	&	99.5	&	100	&	100	\\
agaricus\_lepiota	&	98.1	&	100	&	100	&	100	\\
australian	&	99.8	&	98.9	&	100	&	100	\\
breast\_w	&	97.6	&	97.7	&	99.6	&	99.6	\\
buggyCrx	&	99.2	&	99.6	&	100	&	100	\\
chess	&	99.7	&	100	&	100	&	100	\\
churn	&	98	&	99.1	&	99.9	&	100	\\
clean2	&	99.6	&	99.3	&	99.7	&	99.7	\\
coil2000	&	99.9	&	99.9	&	100	&	100	\\
credit\_a	&	99.6	&	97.5	&	99.9	&	100	\\
credit\_g	&	99.5	&	99.8	&	100	&	100	\\
crx	&	99.4	&	99.4	&	100	&	100	\\
diabetes	&	94.4	&	92.8	&	98	&	98.4	\\
dis	&	97.6	&	99.6	&	99.9	&	100	\\
GAMETES\_1	&	99.7	&	100	&	100	&	100	\\
GAMETES\_2	&	99.3	&	99.9	&	100	&	100	\\
GAMETES\_3	&	99.7	&	100	&	100	&	100	\\
GAMETES\_4	&	99.6	&	100	&	100	&	100	\\
german	&	98.8	&	99.9	&	100	&	100	\\
Hill\_Valley\_without\_noise	&	99.6	&	99.7	&	100	&	100	\\
hypothyroid	&	98.6	&	98.9	&	99.9	&	100	\\
kr\_vs\_kp	&	99.6	&	100	&	100	&	100	\\
magic	&	93.6	&	94.6	&	98.6	&	99.9	\\
mofn\_3\_7\_10	&	97.2	&	99.9	&	100	&	100	\\
monk1	&	97.5	&	98.3	&	99.6	&	99.8	\\
monk2	&	97.2	&	98.2	&	99.8	&	99.9	\\
monk3	&	95.1	&	93.5	&	96.6	&	99.1	\\
mushroom	&	97.8	&	100	&	100	&	100	\\
parity5+5	&	99.5	&	99.9	&	100	&	100	\\
phoneme	&	86.8	&	89.8	&	97.5	&	98.9	\\
pima	&	96.8	&	90.7	&	98.3	&	98.3	\\
profb	&	98.6	&	95.5	&	99.7	&	100	\\
ring	&	98.2	&	99.6	&	99.9	&	100	\\
spambase	&	97.5	&	99.7	&	100	&	100	\\
threeOf9	&	98.6	&	98.9	&	99.8	&	99.9	\\
tic\_tac\_toe	&	99.9	&	99.2	&	100	&	100	\\
tokyo1	&	99.4	&	99.5	&	100	&	100	\\
twonorm	&	86.6	&	99	&	99.7	&	99.9	\\
wdbc	&	98.6	&	99.3	&	99.8	&	100	\\
xd6	&	98.4	&	97	&	99.3	&	100	\\
\midrule
Average	&	97	&	98.6	&	99.6	&	99.9	\\
Standard Deviation	&	3.4	&	1.9	&	0.9	&	0.2	\\

\bottomrule
    \end{tabular}
    \end{adjustbox}
    \captionof{table}{Relative feature exclusion\newline for the RF classifier}
    \label{tab:exclrf}
\end{minipage}
\hfill
\begin{minipage}{0.5\textwidth}
\begin{adjustbox}{max totalsize={0.95\textwidth}{1\textheight},center}
\begin{tabular}{lcccc}
\toprule
Dataset	&	Prox	&	Plaus	&	All CF	&	All	\\
\midrule
adult	&	99.4	&	99.4	&	100	&	100	\\
agaricus\_lepiota	&	97.7	&	97.3	&	99.8	&	100	\\
australian	&	98.4	&	99.9	&	100	&	100	\\
breast\_w	&	99.5	&	99.6	&	99.9	&	100	\\
buggyCrx	&	98.2	&	99.9	&	100	&	100	\\
chess	&	99.5	&	100	&	100	&	100	\\
churn	&	99.4	&	98.9	&	99.9	&	100	\\
clean2	&	99.9	&	99.8	&	100	&	100	\\
coil2000	&	99.6	&	99.7	&	100	&	100	\\
credit\_a	&	98.6	&	99.3	&	100	&	100	\\
credit\_g	&	99.2	&	99.9	&	100	&	100	\\
crx	&	98.2	&	100	&	100	&	100	\\
diabetes	&	89.8	&	95.9	&	97.4	&	99.9	\\
dis	&	99.8	&	100	&	100	&	100	\\
GAMETES\_1	&	99.3	&	100	&	100	&	100	\\
GAMETES\_2	&	99.1	&	99.8	&	99.9	&	100	\\
GAMETES\_3	&	99.7	&	99.9	&	100	&	100	\\
GAMETES\_4	&	99.4	&	100	&	100	&	100	\\
german	&	99	&	100	&	100	&	100	\\
Hill\_Valley\_without\_noise	&	99.1	&	100	&	100	&	100	\\
hypothyroid	&	96	&	99.4	&	99.8	&	100	\\
kr\_vs\_kp	&	99.4	&	100	&	100	&	100	\\
magic	&	89.8	&	96	&	98.5	&	100	\\
mofn\_3\_7\_10	&	98.3	&	99.9	&	100	&	100	\\
monk1	&	95	&	98.4	&	99.4	&	99.9	\\
monk2	&	97.4	&	98.2	&	99.8	&	100	\\
monk3	&	93.3	&	93.8	&	95.6	&	98.8	\\
mushroom	&	97.4	&	96.4	&	99.6	&	100	\\
parity5+5	&	99.7	&	100	&	100	&	100	\\
phoneme	&	92.5	&	92.9	&	98.5	&	99.3	\\
pima	&	93	&	98.6	&	99.4	&	99.5	\\
profb	&	95.6	&	96.4	&	98.8	&	100	\\
ring	&	98.3	&	98.5	&	99.9	&	99.9	\\
spambase	&	99	&	100	&	100	&	100	\\
threeOf9	&	98.3	&	96.7	&	99.5	&	99.8	\\
tic\_tac\_toe	&	99.1	&	96.6	&	99.7	&	99.9	\\
tokyo1	&	89.9	&	100	&	100	&	100	\\
twonorm	&	90.3	&	99.6	&	100	&	100	\\
wdbc	&	88.2	&	99.9	&	100	&	100	\\
xd6	&	97.1	&	94.2	&	98.1	&	100	\\
\midrule
Average	&	97.8	&	98.4	&	99.6	&	99.8	\\
Standard Deviation	&	3	&	2.6	&	0.8	&	0.4	\\
\bottomrule
    \end{tabular}
    \end{adjustbox}
    \captionof{table}{Relative feature exclusion\newline for the ANN classifier}
    \label{tab:exclann}
\end{minipage}

\begin{minipage}{0.5\textwidth}
\begin{adjustbox}{max totalsize={0.95\textwidth}{1\textheight},center}
\begin{tabular}{lcccc}
\toprule
Dataset	&	Prox	&	Plaus	&	All CF	&	All	\\
\midrule
adult	&	40.5	&	48.4	&	59.7	&	74.1	\\
agaricus\_lepiota	&	32.7	&	53	&	62.4	&	65.4	\\
australian	&	45.3	&	57.7	&	63.8	&	74.1	\\
breast\_w	&	56.8	&	85.3	&	87.8	&	88.9	\\
buggyCrx	&	29.3	&	60.8	&	64.4	&	75	\\
chess	&	7.3	&	17.6	&	20.3	&	25.8	\\
churn	&	47.2	&	87.9	&	88.9	&	93.2	\\
clean2	&	14.7	&	92.4	&	92.8	&	100	\\
coil2000	&	17.3	&	31.3	&	39.8	&	41.8	\\
credit\_a	&	41.1	&	57.6	&	62.7	&	75.1	\\
credit\_g	&	27.7	&	55.3	&	59.4	&	65.6	\\
crx	&	27.1	&	59.4	&	64.5	&	74	\\
diabetes	&	57.2	&	89.1	&	91.8	&	93.2	\\
dis	&	17.1	&	24.4	&	24.8	&	49.6	\\
GAMETES\_1	&	17	&	45	&	49.8	&	53	\\
GAMETES\_2	&	15.8	&	39.4	&	44.2	&	67.8	\\
GAMETES\_3	&	16.8	&	39	&	43.6	&	62.3	\\
GAMETES\_4	&	17.4	&	45.5	&	50.6	&	70.6	\\
german	&	24.4	&	57.2	&	60.2	&	65.7	\\
Hill\_Valley\_without\_noise	&	17.4	&	100	&	100	&	100	\\
hypothyroid	&	24.5	&	27.9	&	31.6	&	38.2	\\
kr\_vs\_kp	&	7.9	&	17.6	&	20.5	&	26.5	\\
magic	&	99.9	&	100	&	100	&	100	\\
mofn\_3\_7\_10	&	31.9	&	39.9	&	45.8	&	57.4	\\
monk1	&	36.5	&	42.5	&	49.5	&	63.8	\\
monk2	&	41.1	&	49.4	&	59.8	&	85.8	\\
monk3	&	24.8	&	36.9	&	41.2	&	60.3	\\
mushroom	&	30.6	&	51.8	&	60.6	&	64.7	\\
parity5+5	&	23.8	&	26.6	&	38	&	87.7	\\
phoneme	&	98.4	&	100	&	100	&	100	\\
pima	&	64.8	&	88.2	&	90.9	&	99.9	\\
profb	&	25.2	&	72.1	&	73.6	&	90.1	\\
ring	&	27.9	&	100	&	100	&	100	\\
spambase	&	30.3	&	29.9	&	35.7	&	94	\\
threeOf9	&	25.7	&	25.4	&	34.4	&	56.7	\\
tic\_tac\_toe	&	26.2	&	45.1	&	52.1	&	73.5	\\
tokyo1	&	82.4	&	81.2	&	87.9	&	88	\\
twonorm	&	100	&	100	&	100	&	100	\\
wdbc	&	98.4	&	99.1	&	100	&	100	\\
xd6	&	26.1	&	21.6	&	31.1	&	53.6	\\
\midrule
Average	&	37.4	&	57.5	&	62.1	&	73.9	\\
Standard Deviation	&	25.8	&	27.1	&	25.2	&	21	\\

\bottomrule
    \end{tabular}
    \end{adjustbox}
    \captionof{table}{Relative feature span\newline for the RF classifier}
    \label{tab:spanrf}
\end{minipage}
\hfill
\begin{minipage}{0.5\textwidth}
\begin{adjustbox}{max totalsize={0.95\textwidth}{1\textheight},center}
\begin{tabular}{lcccc}
\toprule
Dataset	&	Prox	&	Plaus	&	All CF	&	All	\\
\midrule
adult	&	28.2	&	55.3	&	61.4	&	68	\\
agaricus\_lepiota	&	23.6	&	52.1	&	57.6	&	60	\\
australian	&	36.2	&	63.8	&	67	&	76.5	\\
breast\_w	&	41.4	&	82.8	&	87.8	&	91.2	\\
buggyCrx	&	24.3	&	61.6	&	64.2	&	65.7	\\
chess	&	8.6	&	17.6	&	21.2	&	27.8	\\
churn	&	54.8	&	88.1	&	90.9	&	93.8	\\
clean2	&	8.4	&	92.6	&	93.1	&	93.1	\\
coil2000	&	17.6	&	30	&	38.3	&	40.2	\\
credit\_a	&	25	&	58.5	&	61.8	&	73.1	\\
credit\_g	&	20.6	&	56.9	&	58.9	&	65.5	\\
crx	&	31.3	&	57.3	&	63.4	&	72.1	\\
diabetes	&	54.8	&	89.6	&	92	&	92.4	\\
dis	&	17.5	&	24.8	&	27.6	&	29.3	\\
GAMETES\_1	&	17.5	&	43.8	&	48.9	&	71.3	\\
GAMETES\_2	&	15.7	&	40.2	&	44.7	&	66.4	\\
GAMETES\_3	&	16.4	&	41.6	&	45.8	&	65.4	\\
GAMETES\_4	&	16.8	&	46.1	&	51	&	69.5	\\
german	&	21.6	&	57.8	&	61.2	&	67.4	\\
Hill\_Valley\_without\_noise	&	21.3	&	100	&	100	&	100	\\
hypothyroid	&	19.8	&	28.7	&	30.9	&	37.9	\\
kr\_vs\_kp	&	9.1	&	17.5	&	21.4	&	28	\\
magic	&	99.7	&	100	&	100	&	100	\\
mofn\_3\_7\_10	&	34.2	&	40.4	&	46.9	&	57	\\
monk1	&	38.7	&	42.4	&	52.1	&	87.5	\\
monk2	&	44.9	&	48.8	&	60	&	89.3	\\
monk3	&	32.1	&	38.5	&	46.2	&	61.3	\\
mushroom	&	26.2	&	50.3	&	56.8	&	62	\\
parity5+5	&	25.1	&	33.2	&	42.4	&	88.9	\\
phoneme	&	97.7	&	99.8	&	99.9	&	100	\\
pima	&	49	&	90.1	&	91.4	&	91.8	\\
profb	&	35.2	&	77.1	&	79.2	&	92.4	\\
ring	&	32.7	&	100	&	100	&	100	\\
spambase	&	29.4	&	34.9	&	38.4	&	40.1	\\
threeOf9	&	26.1	&	33.9	&	42.4	&	61.9	\\
tic\_tac\_toe	&	24.6	&	58.2	&	64.8	&	77.5	\\
tokyo1	&	73.6	&	85	&	85.5	&	88.4	\\
twonorm	&	99.9	&	100	&	100	&	100	\\
wdbc	&	97.8	&	100	&	100	&	100	\\
xd6	&	26.8	&	30.2	&	38.8	&	58.5	\\
\midrule
Average	&	35.6	&	59.2	&	63.3	&	72.8	\\
Standard Deviation	&	24.8	&	26	&	24	&	21.4	\\

\bottomrule
    \end{tabular}
    \end{adjustbox}
    \captionof{table}{Relative feature span\newline for the ANN classifier}
    \label{tab:spanann}
\end{minipage}

Selecting random features that are desired to be in an explanation seems slightly more difficult. The average relative feature span for all counterfactual methods (depicted in \cref{tab:spanrf} and \ref{tab:spanann}) is 62.1\% (63.3\%) for an RF (ANN) classifier, and 73.9\% (72.8\%) if we include Anchors and SHAP. However, there are still datasets that have a relative features span of 100.0\%. Relative feature spans seem to vary tremendously over different datasets, groups of XAI methods, and classifiers. We refer to \cref{sec:question2} for a more detailed examination of the drivers of this counterfactual disagreement.

To conclude, a malicious agent will be easily able to both exclude and include desired features when multiple counterfactual algorithms are available. Especially excluding certain features to hide their influence in the ML model, while still using them to generate predictions, is an almost effortless job. 

\subsubsection{What are the drivers of counterfactual disagreement?}\label{sec:question2}
The disagreement problem makes it easy for malicious agents to exclude or include certain features. While the variation in feature exclusion is minimal, \cref{tab:spanann} and \cref{tab:spanrf} showed that feature exclusion does have some variation. In this section, we investigate what are the drivers of this variation. We investigate whether, the dataset, the counterfactual algorithms, or the classifier cause the variation in counterfactual disagreement. This should help to identify when the possibility of feature disagreement is high. 

\paragraph{Dataset}
\cref{tab:spanann} and \cref{tab:spanrf} show that there is a lot of variance over the different datasets. Overall counterfactual explanations, the relative features span varies from around 20 to 100\% for both classifiers.  

\begin{figure}[hhtp]
    \centering
    \begin{subfigure}[b]{0.45\textwidth}
    \begin{tikzpicture}[scale=0.7]
  \begin{axis}
    [
    ytick={1,2,3,4},
    yticklabels={\texttt{Prox}, \texttt{Plaus}, All CF, All},
    ]
    \addplot+[
    boxplot prepared={
    average=37.4,
      median=27.8,
      upper quartile=42.15,
      lower quartile=22.2,
      upper whisker=100,
      lower whisker=7.3
    },
    ] coordinates {};
    \addplot+[
    boxplot prepared={
    average=57.53,
      median=52.4,
      upper quartile=85.9,
      lower quartile=38.5,
      upper whisker=100,
      lower whisker=17.6
    },
    ] coordinates {
    };
    \addplot+[
    boxplot prepared={
    average=62.1,
      median=60,
      upper quartile=88.15,
      lower quartile=43,
      upper whisker=100,
      lower whisker=20.3
    },
    ] coordinates {};
     \addplot+[
    boxplot prepared={
    average=73.8,
      median=74.05,
      upper quartile=93.2,
      lower quartile=61.8,
      upper whisker=100,
      lower whisker=25.8
    },
    ] coordinates {};
  \end{axis}
\end{tikzpicture}
\caption{RF}\label{fig:boxplotRF}
    \end{subfigure}
    \hfill
    \begin{subfigure}[b]{0.45\textwidth}
         \begin{tikzpicture}[scale=0.7]
  \begin{axis}
    [
    ytick={1,2,3,4},
    yticklabels={\texttt{Prox}, \texttt{Plaus}, All CF, All},
    ]
    \addplot+[
    boxplot prepared={
    average=35.6,
      median=26.5,
      upper quartile=39.3,
      lower quartile=20.4,
      upper whisker=99.9,
      lower whisker=8.4
    },
    ] coordinates {};
    \addplot+[
    boxplot prepared={
    average=59.23,
      median=56.1,
      upper quartile=85.7,
      lower quartile=39.7,
      upper whisker=100,
      lower whisker=17.5
    },
    ] coordinates {
    };
    \addplot+[
    boxplot prepared={
    average=63.3,
      median=60.6,
      upper quartile=88.57,
      lower quartile=45.5,
      upper whisker=100,
      lower whisker=21.8
    },
    ] coordinates {};
     \addplot+[
    boxplot prepared={
    average=72.78,
      median=71.7,
      upper quartile=91.95,
      lower quartile=61.75,
      upper whisker=100,
      lower whisker=27.8
    },
    ] coordinates {};
  \end{axis}
\end{tikzpicture}
\caption{ANN}\label{fig:boxplotANN}
\end{subfigure}
    \caption{Box-plots relative feature span for different algorithm groups for the ANN and RF classifiers}
    \label{fig:boxplot}
\end{figure}
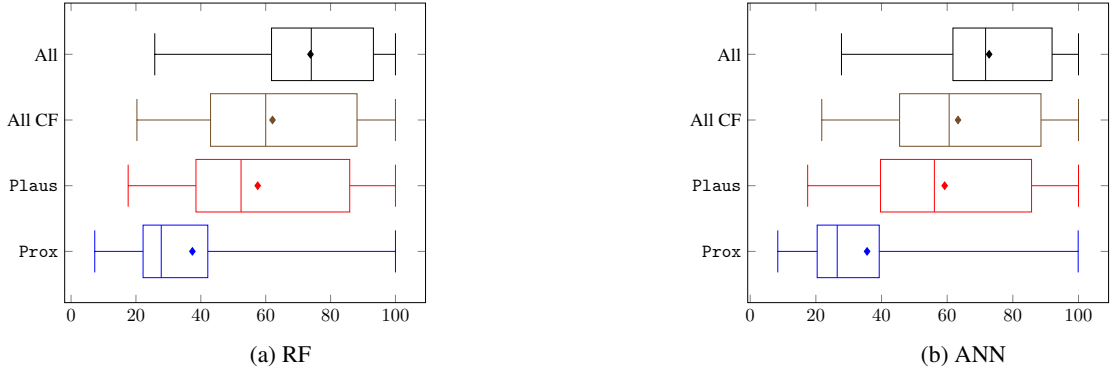

\paragraph{Counterfactual algorithms}
As shown in \cref{tab:tableAlgoOverview}, the counterfactual algorithms can be divided into two groups \texttt{plaus} and \texttt{prox}. Those that optimize for plausibility, and those that only optimize for plausibility. It might be that the existing disagreement only originates from the disagreement between these groups and not from the disagreement within these groups. To verify this, we also calculated the relative feature span within these groups in \cref{tab:spanann} and \ref{tab:spanrf} and a noticeable difference can be seen. The span within \texttt{plaus} is 20.1 to 23.6\% higher compared to the span within the \texttt{prox} group. \Cref{fig:boxplot} stresses the difference between the two groups visually by the use of box plots. The center of gravity for the \texttt{prox} group is not only lower but also less broad, compared to the \texttt{plaus} group, meaning that even though the variance stretches over the entire x-axis, the gross of the relative feature spans for this group lies between 22 and 42\% (20 and 39\%) for the RF (ANN) classifier. In contrast, the relative feature span for the \texttt{plaus} group lies mainly between 39 and 86\% (40 and 86\%) for the RF (ANN) classifier.
This difference can simply be explained by the difference in sparsity between both groups as seen in \cref{tab:averagespars}. Sparsity refers to the number of features in a counterfactual explanation. Optimizing for proximity (or sparsity directly) has a direct effect on this number of features in the explanations. Therefore, the average sparsity of the \texttt{prox} group is much lower compared to that of the \texttt{plaus} group. Consequently, having fewer features on average in each explanation also results in a lower relative feature span for this group. Furthermore, the \texttt{plaus} group seems to account for most of the feature span of all counterfactual explanations. The difference between the \texttt{plaus} group and the group of all counterfactual explanations is less than 5\% for both classifiers.

To first have a grasp of how similar the explanations are between different counterfactual algorithms, we examine the pairwise scaled L0 distances between the counterfactual instances of counterfactual methods. This metric counts the number of features that two explanations have in common and divide this number by the total number of features in the dataset.

\begin{equation}
\label{eq:L0}
\textrm{L0 distance}_{ab} = \frac{|E_a \cap E_b|}{F_D}
\end{equation}

In order to quantify the \emph{pairwise} disagreement amongst counterfactual explanations, Anchors, and SHAP, we first introduce a new measure called feature disagreement in \cref{eq:FeatureAgreement}. This measure is similar to the feature agreement metric introduced by \textcite{krishna2022disagreement} but adapted to the variable explanation sizes of which counterfactuals consist.

\begin{equation}
\label{eq:FeatureAgreement}
\textrm{Feature disagreement}_{ab} = \frac{|E_a \setminus E_b|}{|E_a|}
\end{equation}

When comparing the explanations $E_a$ and $E_b$ of two methods $A$ and $B$, the feature disagreement of $A$ with $B$ is equal to the size of the relative complement of $E_b$ in $E_b$ divided by the size of $E_a$. It measures the relative number of features that are in $E_a$ but not in $E_b$. When the feature disagreement equals 1, none of the features of $E_a$ are also in $E_b$.

It could be argued that the Jaccard distance can be used to quantify the pairwise disagreement amongst counterfactual methods, this being an existing metric. However, since the feature disagreement metric contains more information regarding the \emph{direction} of the disagreement, we continue with this metric and refer to Appendix A for the Jaccard distance analysis. 

When revisiting the example in \cref{sec:example}, and once again assuming the user only has the first two counterfactual explanation algorithms, CFproto and WIT, available. The pairwise scaled L0 distance between CFproto and WIT equals $\frac{E_{CFproto} \cap E_{WIT}}{F_{Adult}}$ or 35.7\%. The feature disagreement between CFproto and WIT equals $\frac{|E_{CFproto} \setminus E_{WIT}|}{|E_{CFproto}|}$ or 50\%. CFproto has 50\% unique features with respect to WIT. Vice versa, the feature disagreement between WIT and CFproto equals $\frac{|E_{WIT} \setminus E_{CFproto}|}{|E_{WIT}|}$ or 28.6\%. WIT has 28.6\% unique features with respect to CFproto. 

\begin{figure}[hhtp]
\begin{subfigure}[b]{0.5\textwidth}
   \centering
    \includegraphics[scale=0.45]{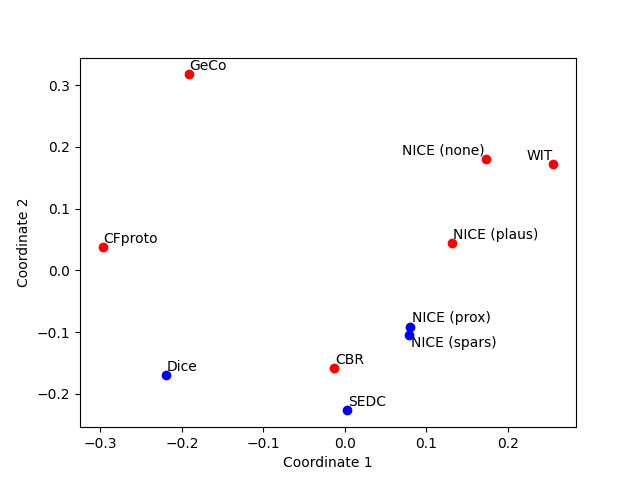}
    \caption{ANN}
    \label{fig:distANNL0}
\end{subfigure}
\hfill
\begin{subfigure}[b]{0.5\textwidth}
  \centering
    \includegraphics[scale=0.45]{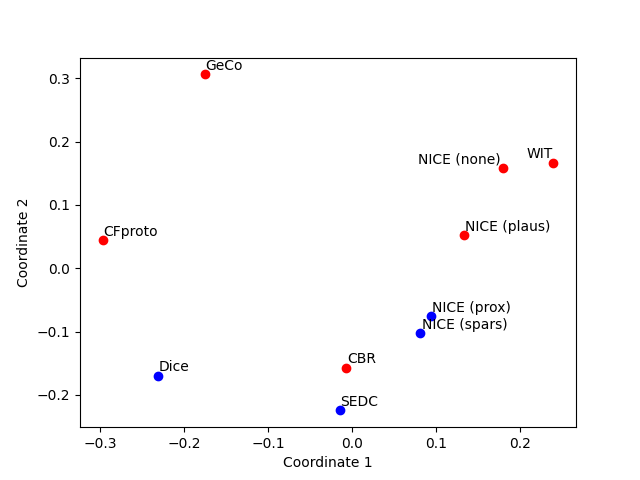}
    \caption{RF}
    \label{fig:distRFL0}
\end{subfigure}
\caption{Multidimensional scaling (MDS) for the L0 distance between the counterfactual explanation methods (\textcolor{blue}{Blue:} \texttt{Prox}, \textcolor{red}{Red:} \texttt{Plaus})} \label{fig:distanceL0}
\end{figure}

The pairwise scaled L0 distances are shown in \cref{tab:L0} in Appendix A. We visualized these distances in a 2D-plot by using multidimensional scaling (MDS) in \cref{fig:distRFL0} and \cref{fig:distANNL0}. Note that the numbers on the x- and y-axis of these figures have no translatable meaning, only the relative Euclidean distances between two points are meaningful. The closer two points lie together, the more similar the resulting counterfactual instances of each method are. NICE(prox) and NICE(spars) optimize for very similar metrics with the same optimization method and therefore result in very similar counterfactual instances. The same can be said for NICE(none) and WIT. Both these methods use real instances from the training set in their explanations, and those instances seem to be quite close to each other. Surprisingly, SEDC and CBR are similar as well.
GeCo provides counterfactual instances that are the farthest away from all other methods. This might be because GeCo's explanations contain many features in general. \cref{tab:averagespars} shows, that GeCo on average uses around 74\% of all features in a single counterfactual explanation.

Which algorithms disagree the most with others, while taking into account the number of features present in their explanations, can be identified by looking at the pairwise feature disagreement. \Cref{tab:disagreement} shows that most post-hoc explanation methods have a high number of features that are not present in CBR, SEDC or SHAP (the darkest columns in \cref{tab:disagreement}) even though the sparsity of these methods is not necessarily low (see \cref{tab:averagespars}).  Overall GeCo is the counterfactual algorithm that generates the most features that are not available in the explanations of other algorithms. Once again, this can be largely attributed to the fact that GeCo has the highest sparsity, meaning that it has the most features in its generated explanations. 

The high feature disagreement of SHAP and Anchors with counterfactual explanations confirms that the disagreement between different post-hoc explanation methods is larger than the disagreement within counterfactual explanations. 

Lastly \cref{tab:groupdis}, presents the average pairwise disagreement between a counterfactual method presented in the first column and the other counterfactual algorithms within the same (intra) and between (inter) group(s). For example, the intra group average for CBR (a member of the \texttt{plaus} group) equals the average of the relative feature disagreements between CBR and the other members of the \texttt{plaus} group: CFproto, WIT, GeCo, NICE(none) and NICE(plaus). On the other hand, the inter group average, amounts the average of the relative feature disagreements between CBR and the members of the \texttt{prox} group: DiCe, NICE(prox), NICE(spars), and SEDC. 
It is clear that for the \texttt{plaus} group, the intra group averages are significantly lower compared to the inter group averages. In contrast, for the \texttt{prox} group the inter group averages are lower. Since counterfactual algorithms of the \texttt{prox} group generate explanations with less features compared to the other group, chances of disagreement are higher. Vice Versa, it is easier to find agreement when a lot of features are present in the explanantions, which is the case in the \texttt{plaus} group.

\begin{table}
 \begin{adjustbox}{max totalsize={1\textwidth}{0.65\textheight},center}
\centering
\pgfplotstabletypeset[color cells]{
RF,CBR,CFproto,WIT,GeCo,NICE(none),NICE(plaus),DiCE,NICE(prox),NICE(spars),SEDC,Anchors,SHAP,Average
CBR,0.0,37.0,14.3,28.9,15.1,20.3,35.4,25.2,25.0,39.1,12.1,43.3,24.6
CFproto,72.0,0.0,20.7,30.3,20.8,33.6,44.5,45.1,47.8,61.5,34.7,57.4,39.0
WIT,81.6,56.0,0.0,40.3,4.6,25.8,64.1,48.8,53.5,78.4,57.6,79.7,49.2
GeCo,87.3,63.4,35.7,0.0,36.7,51.0,70.1,69.0,72.7,82.0,63.9,77.9,59.1
NICE(none),81.9,55.7,4.0,40.6,0.0,22.6,64.0,47.2,52.0,78.7,58.0,80.2,48.7
NICE(plaus),78.0,54.4,3.5,41.1,0.0,0.0,60.2,37.4,40.2,70.4,50.3,80.2,43.0
DiCE,80.3,52.0,32.5,43.5,32.4,43.4,0.0,55.3,59.4,71.4,48.3,69.7,49.0
NICE(prox),75.2,51.3,3.0,42.1,0.0,14.3,54.2,0.0,17.2,66.7,41.5,81.9,37.3
NICE(spars),72.1,50.9,3.3,43.3,0.0,11.8,53.5,7.7,0.0,63.0,37.2,80.9,35.3
SEDC,63.7,45.2,20.2,35.1,21.3,28.1,40.4,33.2,34.3,0.0,21.2,58.2,33.4
Anchors,80.4,58.0,33.9,50.1,35.7,45.3,55.4,53.4,55.9,67.7,0.0,79.0,51.2
SHAP,89.7,72.0,57.7,59.6,59.9,67.7,73.3,78.4,79.1,82.3,72.1,0.0,66.0
}
\end{adjustbox}

\vspace{5mm}
\begin{adjustbox}{max totalsize={1\textwidth}{0.65\textheight},center}
\pgfplotstabletypeset[color cells]{
ANN ,CBR,CFproto,WIT,GeCo,NICE(none),NICE(plaus),DiCE,NICE(prox),NICE(spars),SEDC,Anchors,SHAP,Average
CBR,0,39.7,17.1,30.9,17.9,25.7,43,32.7,32.7,47.5,21.9,59.2,30.7
CFproto,74.3,0,24.8,36.9,24.7,39.6,51.6,54.6,56.7,66.6,42.8,75.3,45.7
WIT,80.8,53.5,0,38.9,4.3,29.9,67.3,55.7,57.4,80.2,59.8,87.3,51.3
GeCo,85.7,56.4,33.9,0,34.6,53.1,65.8,73.7,75.4,81.7,66.4,84.6,59.3
NICE(none),80.8,53,3.3,38.8,0,26.9,67,54.3,56.1,80.2,59.9,87.8,50.7
NICE(plaus),75.9,52.2,3.1,39.3,0,0,65.3,40.1,40,70.9,50.9,86.8,43.7
DiCE,82,51,36.8,44.2,36.5,51.8,0,65.7,67.5,75.7,54.8,82.7,54.1
NICE(prox),70.2,46.7,3.1,41.1,0,12.8,58.6,0,10.7,63.9,38.8,87,36.1
NICE(spars),68.4,48,3.5,41.8,0,10.6,59,7,0,62.2,37.1,86.3,35.3
SEDC,66.1,44.8,24,36.4,24.5,32.9,50.3,40.5,41.1,0,29.4,71,38.4
Anchors,77.8,54.5,34.2,46.2,35.3,48.3,60.7,57.2,58.8,69.7,0,86.9,52.5
SHAP,76,62.2,52.4,50.2,54.8,62,61.6,68.8,69.1,70.6,58.6,0,57.2
}
\end{adjustbox}
\caption{Relative feature disagreement for the RF and ANN classifier}\label{tab:disagreement}
\end{table}

\begin{table}[hhtp]
 \begin{adjustbox}{max totalsize={1\textwidth}{0.65\textheight},center}
\centering
\pgfplotstabletypeset[color cells]{
Classifier,CBR,CFproto,WIT,GeCo,NICE(none),NICE(plaus),DiCE,NICE(prox),NICE(spars),SEDC
ANN, 44.1,	50.2,	45.4,	73.6,	44.7,	30.1,32.4,	14.9,	13.9,	31.1
RF, 55.1,	57.9,	44.8,	74.0,	44.4,	32.6,	40.4,	18.7,	15.7,	38.3
}
\end{adjustbox}
\caption{The average sparsity for the RF and ANN classifier}\label{tab:averagespars}
\end{table}

\begin{table}[hhtp]
    \centering
    \begin{adjustbox}{max totalsize={0.55\textwidth}{1\textheight},center}
    \begin{tabular}{lcc|cc}
    \toprule

&\multicolumn{2}{c|}{RF}& \multicolumn{2}{c}{ANN}\\

&Intra group&	Inter group &Intra group&	Inter group\\
\midrule
CBR	&	23.1	&	31.2	&	26.2	&	39		\\
CFproto	&	35.5	&	49.7	&	40.1	&	57.4		\\
WIT	&	41.7	&	61.2	&	41.5	&	65.2		\\
GeCo	&	54.8	&	73.5	&	52.7	&	74.1		\\
NICE(none)	&	41.0	&	60.5	&	40.5	&	64.4		\\
NICE(plaus)	&	35.4	&	52.1	&	34.1	&	54.1		\\
\midrule
DiCE	&	62.0	&	47.4	&	69.6	&	50.4	\\
NICE(prox)	&	46.0	&	31.0	&	44.4	&	29.0		\\
NICE(spars)	&	41.4	&	30.2	&	42.7	&	28.7	\\
SEDC	&	36.0	&	35.6	&	44.0	&	38.1	\\

\bottomrule
    \end{tabular}
    \end{adjustbox}
    \caption{Intra and inter group relative feature disagreement}
    \label{tab:groupdis}
\end{table}

\paragraph{Classifier}
Surprisingly, the used classifier does not have a critical influence on the size of the disagreement problem. For each dataset the difference between the average relative feature span between both classifiers is minimal. Moreover, we calculated the correlation between both classifiers, which is more than 99\%. The same conclusion can be drawn from the L0 distances in \cref{fig:distanceL0} between the counterfactual instances or the relative feature disagreements in \cref{tab:disagreement}. Both metrics show little variation between the RF and ANN.

In conclusion, both the dataset and the group of counterfactual algorithms determine the variation in disagreement metrics. Perhaps more surprisingly, we find that the classifier has a small to no influence on the results and variation obtained.

\section{Conclusion and future research}\label{sec:conclusion}
In our large-scale empirical analyses, on 40 datasets, yielding over 192,0000 explanations generated, we confirm the existence of the disagreement problem amongst different counterfactual explanation algorithms. If a malicious agent has the option to choose between the 10 counterfactual algorithms examined in our experiments, it will be very easy to exclude features of their choice in an explanation. Including a feature of choice, is slightly more difficult, but still in many cases the relative feature span is 100\%, giving the decision maker the full choice to include a certain feature.

Moreover, we conclude that the size of the disagreement problem is highly dependent on the dataset and counterfactual methods used and not so much on the classifier used.
However, we want to stress again that in contrast to other post-hoc explanation methods, disagreement between counterfactual explanations does not mean any explanation is wrong. On one hand, a counterfactual explanation cannot be wrong, as the suggested feature changes will by definition lead to a class change. It can, however, be that these suggested changes are not useful to certain stakeholders. Therefore, situations with high disagreement between counterfactual explanations signal instead that one single explanation fails to capture the full complexity of a decision made by an ML model.

 By proving the existence of the disagreement problem amongst counterfactual explanation methods, we demonstrate the potential rise of ethical issues. Especially in an adversarial context, where the goals of the stakeholders are not aligned, these ethical issues occur when users are able to choose which explanations are used, giving them a lot of power. To avoid the occurrence of moral issues, ideally, this power should be in the hands of the decision subject, as they carry the, possibly life-changing, consequences of the decision. Giving the explanatory decision power to the decision subject, can in turn create new issues. Decision subjects are often not familiar with model outputs. Offering them multiple explanations goes against the understandability of human nature and can become overwhelming. In this case, the XAI method would fail its primary goal: making AI decisions interpretable to humans.

To avoid ethical issues, decision-makers should be transparent about the transparency given by XAI. They should be open about every step of their decision-making process, which explanation method, which metric is optimized, and which individual explanation is used. All these steps should be motivated to ensure that explanations are never used to justify biased decision-making. We argue laws or policies such as GDPR should take the above-mentioned into account for the necessary future policy development concerning the use of (X)AI. Quality explanations are currently ill-defined. Explanation methods can therefore be freely chosen without transparency, and consequently, the door for unethical behavior is open. Post-hoc explanation methods can currently be used to justify decisions by unfair or even discriminating models. What legislators can do, however, is force \emph{transparency in transparency}: decision-makers should be transparent about their transparency and thus, explain how their explanations are generated. This way, malicious users can be held accountable.

\section*{Acknowledgements}
Dr. Lissa Melis was supported by a Fellowship of the Belgian American Educational Foundation (BAEF) and the President's Postdoctoral Fellowship Program (PPFP).

\clearpage
\printbibliography
\clearpage
\appendix
\section*{Appendix A: Scaled L0 distance} \label{sec:L0}
\renewcommand{\thefigure}{A\arabic{figure}}
\renewcommand{\thetable}{A\arabic{table}}
\setcounter{figure}{0}
\setcounter{table}{0}

\begin{table}[h]
 \begin{adjustbox}{max totalsize={1\textwidth}{0.65\textheight},center}
\centering
\pgfplotstabletypeset[color cells]{
RF,CBR,Cfproto,WIT,GeCo,NICE(none),NICE(plaus),DiCE,NICE(prox),NICE(spars),SEDC
CBR,0,38.5,43.1,43.7,42.8,25.6,30.1,15.7,13.4,13.6
Cfproto,38.5,0,46.8,49.1,46.4,42.8,37.1,39.1,38.9,39.3
WIT,43.1,46.8,0,46.3,12.9,21.4,47.7,30.7,32.8,44.9
GeCo,43.7,49.1,46.3,0,46.7,45.3,45.7,44.6,44.3,43
NICE(none),42.8,46.4,12.9,46.7,0,11.8,47.3,25.7,28.6,44.8
NICE(plaus),25.6,42.8,21.4,45.3,11.8,0,41.7,18.1,19.3,30.5
DiCE,30.1,37.1,47.7,45.7,47.3,41.7,0,33.7,32.8,31.2
NICE(prox),15.7,39.1,30.7,44.6,25.7,18.1,33.7,0,5.1,20.9
NICE(spars),13.4,38.9,32.8,44.3,28.6,19.3,32.8,5.1,0,18.3
SEDC,13.6,39.3,44.9,43,44.8,30.5,31.2,20.9,18.3,0

}
\end{adjustbox}

\vspace{5mm}
\begin{adjustbox}{max totalsize={1\textwidth}{0.65\textheight},center}
\pgfplotstabletypeset[color cells]{
ANN,CBR,Cfproto,WIT,GeCo,NICE(none),NICE(plaus),DiCE,NICE(prox),NICE(spars),SEDC
CBR,0,39,43.1,45.7,42.7,28.5,29.6,15.1,14,14.7
Cfproto,39,0,47.5,49.7,47.2,43.3,36.7,38.9,38.9,38.6
WIT,43.1,47.5,0,48,16.2,25.7,48.2,34.8,35.4,45.8
GeCo,45.7,49.7,48,0,48.2,47,46.8,46,46,46.4
NICE(none),42.7,47.2,16.2,48.2,0,14.7,47.5,29.8,30.8,45.6
NICE(plaus),28.5,43.3,25.7,47,14.7,0,40.5,18,18.1,29.1
DiCE,29.6,36.7,48.2,46.8,47.5,40.5,0,31.5,31.1,30.8
NICE(prox),15.1,38.9,34.8,46,29.8,18,31.5,0,2.7,17.3
NICE(spars),14,38.9,35.4,46,30.8,18.1,31.1,2.7,0,16.5
SEDC,14.7,38.6,45.8,46.4,45.6,29.1,30.8,17.3,16.5,0

}
\end{adjustbox}
\caption{Scaled L0 distance for the RF and ANN classifier}\label{tab:L0}
\end{table}

\newpage
\section*{Appendix B: Jaccard distance} \label{sec:jaccard}
\renewcommand{\thefigure}{B\arabic{figure}}
\renewcommand{\thetable}{B\arabic{table}}
\renewcommand{\theequation}{B\arabic{equation}}
\setcounter{figure}{0}
\setcounter{table}{0}
\setcounter{equation}{0}

\Cref{eq:jaccard} counts the number of features two explanations have in common as the numerator and the union of both explanations in the denominator. The Jaccard distance always lies between 0 and 1. The complement, 1 minus the Jaccard distance, gives an indication of the dissimilarity of two explanations (\cref{eq:jaccarddiss}). 

\begin{equation}
\label{eq:jaccard}
\textrm{Jaccard distance or similarity}_{ab} = \frac{|E_a \cap E_b|}{|E_a \cup E_b|}
\end{equation}

\begin{equation}
\label{eq:jaccarddiss}
\textrm{Jaccard dissimilarity}_{ab} = 1 - \frac{|E_a \cap E_b|}{|E_a \cup E_b|}
\end{equation}
\vspace{2mm}

\Cref{tab:jaccardinv} shows the pairwise inverse Jaccard distance or dissimilarity.

\begin{table}[h]
 \begin{adjustbox}{max totalsize={1\textwidth}{0.65\textheight},center}
\centering
\pgfplotstabletypeset[color cells]{
RF,CBR,CFproto,WIT,GeCo,NICE(none),NICE(plaus),DiCE,NICE(prox),NICE(spars),SEDC,Anchors,SHAP,Average
CBR,0,89.7,82.4,92.3,82.7,80.5,90.3,79.1,77.2,82.7,83.7,95.2,78
CFproto,89.7,0,58.7,72.5,58.7,66.6,74.5,74.2,76.7,84,80.9,90.6,68.9
WIT,82.4,58.7,0,50.6,6.7,27.1,68.8,49.2,53.8,79.7,74.6,85.4,53.1
GeCo,92.3,72.5,50.6,0,51.2,62.3,77.9,76.7,79.8,87.8,80,84.4,68
NICE(none),82.7,58.7,6.7,51.2,0,22.6,68.6,47.2,52,80,75,85.9,52.6
NICE(plaus),80.5,66.6,27.1,62.3,22.6,0,71.5,40.4,41.1,72.6,74.8,87.6,53.9
DiCE,90.3,74.5,68.8,77.9,68.6,71.5,0,74.7,76.9,83.2,79.4,90.9,71.4
NICE(prox),79.1,74.2,49.2,76.7,47.2,40.4,74.7,0,18.8,70.1,71.5,91.5,57.8
NICE(spars),77.2,76.7,53.8,79.8,52,41.1,76.9,18.8,0,67.4,69.8,91.2,58.7
SEDC,82.7,84,79.7,87.8,80,72.6,83.2,70.1,67.4,0,76.4,92.3,73
Anchors,83.7,80.9,74.6,80,75,74.8,79.4,71.5,69.8,76.4,0,88.9,71.3
SHAP,95.2,90.6,85.4,84.4,85.9,87.6,90.9,91.5,91.2,92.3,88.9,0,82
}
\end{adjustbox}

\vspace{5mm}
\begin{adjustbox}{max totalsize={1\textwidth}{0.65\textheight},center}
\pgfplotstabletypeset[color cells]{
ANN,CBR,CFproto,WIT,GeCo,NICE(none),NICE(plaus),DiCE,NICE(prox),NICE(spars),SEDC,Anchors,SHAP,Average
CBR,0,88.4,81.6,90.8,81.6,78.5,90.2,75.2,74,83.1,82,94.5,76.7
CFproto,88.4,0,56.5,66.5,56.2,65.8,73.8,75,76.8,83.7,81.1,93.5,68.1
WIT,81.6,56.5,0,47.7,6.1,30.9,70.7,55.9,57.6,81.4,76.4,91.1,54.7
GeCo,90.8,66.5,47.7,0,48.1,63.4,72.9,80.1,81.6,87.3,81.8,90.4,67.6
NICE(none),81.6,56.2,6.1,48.1,0,26.9,70.5,54.3,56.1,81.5,76.6,91.5,54.1
NICE(plaus),78.5,65.8,30.9,63.4,26.9,0,76.2,42.2,40.9,73.4,75.9,91.5,55.5
DiCE,90.2,73.8,70.7,72.9,70.5,76.2,0,78.6,79.5,86.4,80.5,93,72.7
NICE(prox),75.2,75,55.9,80.1,54.3,42.2,78.6,0,12.2,69.3,70.2,92,58.8
NICE(spars),74,76.8,57.6,81.6,56.1,40.9,79.5,12.2,0,68.1,70.3,91.6,59.1
SEDC,83.1,83.7,81.4,87.3,81.5,73.4,86.4,69.3,68.1,0,78.4,92.1,73.7
Anchors,82,81.1,76.4,81.8,76.6,75.9,80.5,70.2,70.3,78.4,0,90.2,72
SHAP,94.5,93.5,91.1,90.4,91.5,91.5,93,92,91.6,92.1,90.2,0,84.3
}
\end{adjustbox}
\caption{Inverse Jaccard distance (dissimilarity) for the RF and ANN classifier}\label{tab:jaccardinv}
\end{table}

\end{document}